\documentclass{article}
\usepackage{spconf,amsmath,graphicx}
\usepackage{epstopdf}
\usepackage{array}
\usepackage{graphicx}
\usepackage{multirow}
\usepackage{color, soul}
\usepackage{cite}
\usepackage{url}
\usepackage{hyperref}
\usepackage{booktabs}
\usepackage{amsmath}

\usepackage{amssymb}

\title{GraphSpeech: syntax-aware graph attention network \\ for Neural Speech Synthesis}

\name{Rui Liu$^{ 1}$ \thanks{\textbf{Speech samples:} \textcolor{magenta}{\href{https://ttslr.github.io/GraphSpeech/}{https://ttslr.github.io/GraphSpeech/}}}, Berrak Sisman$^{ 2}$, Haizhou Li$^{ 1}$}
\address{ $^1$ National University of Singapore  $^2$ Singapore University of Technology and Design (SUTD) \\
\small{ \{r.liu,berraksisman\}@u.nus.edu, haizhou.li@nus.edu.sg}
}

\begin{document}
\maketitle
\begin{abstract}
Attention-based end-to-end text-to-speech synthesis (TTS) is superior to conventional statistical methods in many ways. Transformer-based TTS is one of such successful implementations. While Transformer TTS models the speech frame sequence well with a self-attention mechanism, it does not associate input text with output utterances from a syntactic point of view at sentence level.
We propose a novel neural TTS model, denoted as \textit{GraphSpeech}, that is formulated under graph neural network framework. \textit{GraphSpeech} encodes explicitly the syntactic relation of input lexical tokens in a sentence, and incorporates such information to derive syntactically motivated character embeddings for TTS attention mechanism. Experiments show that \textit{GraphSpeech} consistently outperforms the Transformer TTS baseline in terms of spectrum and prosody rendering of utterances. 

\end{abstract}
\vspace{-1mm}
%


\vspace{-4mm}
\section{Introduction}
\label{sec:intro}
\vspace{-2mm}

Text-to-speech (TTS) seeks to synthesize human-like natural sounding voice for input text~\cite{taylor2009text,zen2009statistical}. The recent advances have enabled many applications~\cite{sisman2020overview} such as smart voice assistants, dubbing of movies and games, online education, and smart home. In the recent past, concatenative \cite{hunt1996unit} and statistical parametric speech synthesis \cite{ze2013statistical}
systems were the mainstream techniques. We note that both of these techniques have complex pipelines including front-end model, duration model and acoustic model.

With the advent of deep learning, end-to-end generative TTS models simplify the synthesis pipeline with a single neural network. Tacotron-based neural TTS \cite{wang2017tacotron,shen2018natural} and its variants \cite{liu2020teacher,Liu2020spl,liu2020expressive}
are such examples. In these techniques, the key idea is to integrate the conventional TTS pipeline into a unified encoder-decoder network  and to learn the mapping directly from the $<$text, wav$>$ pair.  Tacotron is a successful encoder-decoder implementation based on recurrent neural networks (RNN), such as LSTM \cite{2018interspeech,2018coling}
and GRU \cite{cho2014learning}. However, the recurrent nature inherently limits the possibility of parallel computing in both training and inference.

Self-attention network (SAN) represents another type of encoder-decoder implementation that achieves efficient parallel training with decent performance in machine translation ~\cite{vaswani2017attention}. This model allows for parallel computation as implemented in Transformer TTS~\cite{2019transformerTTS}. Another benefit of SAN is to function with intra-attention~\cite{vaswani2017attention,yang2020localness}
, which has a shorter path to model long distance context. 
Despite the progress~\cite{2019transformerTTS},
Transformer TTS doesn't explicitly associate input text with output utterances from syntactic point of view at sentence level, which is proven useful in speaking style and prosody modeling~\cite{xiao2020improving,liu2021taslp,aso2020acoustic,sun2020graphtts,zhou2020seen}. As a result, the rendering of utterance is adversely affected especially for long sentences. 

Graph neural networks (GNNs) are connectionist models that capture the dependence of graphs via message passing between the nodes of graphs. In graph-to-sequence learning~\cite{ScarselliGTHM09,CohnHB18},
we can frame the self-attention network in the framework of graph neural network, where the token sequence is considered as an unlabeled fully-connected graph (each token as a node), and  the self-attention mechanism is considered as a specific message-passing scheme. Inspired by this, we propose a novel neural TTS model, denoted as \textit{GraphSpeech}, which takes two novel encoding module: Relation Encoder and Graph Encoder. The relation encoder extracts the syntax tree from the input text; then transforms the syntax tree into graph structure; and encode the relationship between any two tokens in the graph. The graph encoder presents the syntax-aware graph attention mechanism, which not only focuses on tokens but also their relation, thus benefit from the knowledge about sentence structure.

The main contributions of this paper are listed as follows:
1) we propose a novel neural TTS architecture, denoted as \textit{GraphSpeech}; 2) we formulate the extension from syntax tree to syntax graph, and character-level relation encoding; 3) we propose a syntax-aware graph attention mechanism, which incorporates linguistic knowledge into the attention calculation; and 4) \textit{GraphSpeech} outperforms the state-of-the-art Transformer TTS in both spectrum and prosody modeling. To our best knowledge, this is the first implementation of syntactically-informed attention mechanism with a graph neural network perspective in Transformer TTS. 

This paper is organized as follows: In Section 2, we re-visit the Transformer TTS framework that serves as a baseline reference. In Section 3, we study the proposed \textit{GraphSpeech}. In Section 4, we report the evaluation results. We conclude the paper in Section 5. 

\vspace{-4mm}
\section{Transformer TTS}
\label{sec:baseline}
\vspace{-2mm}
Transformer TTS~\cite{2019transformerTTS} is a neural TTS model based on self-attention network (SAN)~\cite{vaswani2017attention} with an encoder-decoder architecture which enables parallel training and learning long-distance dependency. 
SAN-based encoder has multi-head self-attention that directly builds the long time dependency between any two tokens. Given a token sequence $x_{1:n}$ of an input text, for each attention head, the attention score between the token $x_{i}$ and the token $x_{j}$ is simply the dot-product of their query vector and key vector, as given below:
\vspace{-2mm}
\begin{equation}
s_{i j}  =f\left(x_{i}, x_{j}\right) =x_{i} W_{q}^{T} W_{k} x_{j}
\label{eq:att0}
\vspace{-3mm}
\end{equation}



The final attention output \textit{attn} is scaled and normalized by a softmax function as follows:
\vspace{-2mm}

\begin{equation}
\vspace{-3mm}
a t t n=\sum_{i=1}^{n} a_{i} W_{v} x_{i}
\end{equation}

\begin{equation}
a_{i}=\frac{\exp (s_{ij} / \sqrt{d})}{\sum_{j=1}^{n} \exp ( s_{ij} / \sqrt{d} )}
\vspace{-2mm}
\end{equation}
where $\sqrt{d}$ is the scaling factor with $d$ being the dimensionality of layer states, and
$W_{q}$, $W_{k}$, and $W_{v}$ are  trainable matrices. Lastly, the outputs of all attention heads are concatenated and projected to obtain the final attention values.





While SAN-encoder builds the long-range dependency between any two tokens to model the global context, it doesn't model the complex syntactic dependency of words within a sentence~ \cite{zhang2020aaai,CaiL20}, which is usually described by a tree structure or a graph. 
We argue that a mechanism that represents the deep syntactic relationship is required to model the association between input text and output utterances. Next we study a novel \textit{GraphSpeech} model in Section \ref{sec:model}.

\begin{figure}[tb]
\vspace{-5mm}
\centering
\centerline{\includegraphics[width=1.03\linewidth]{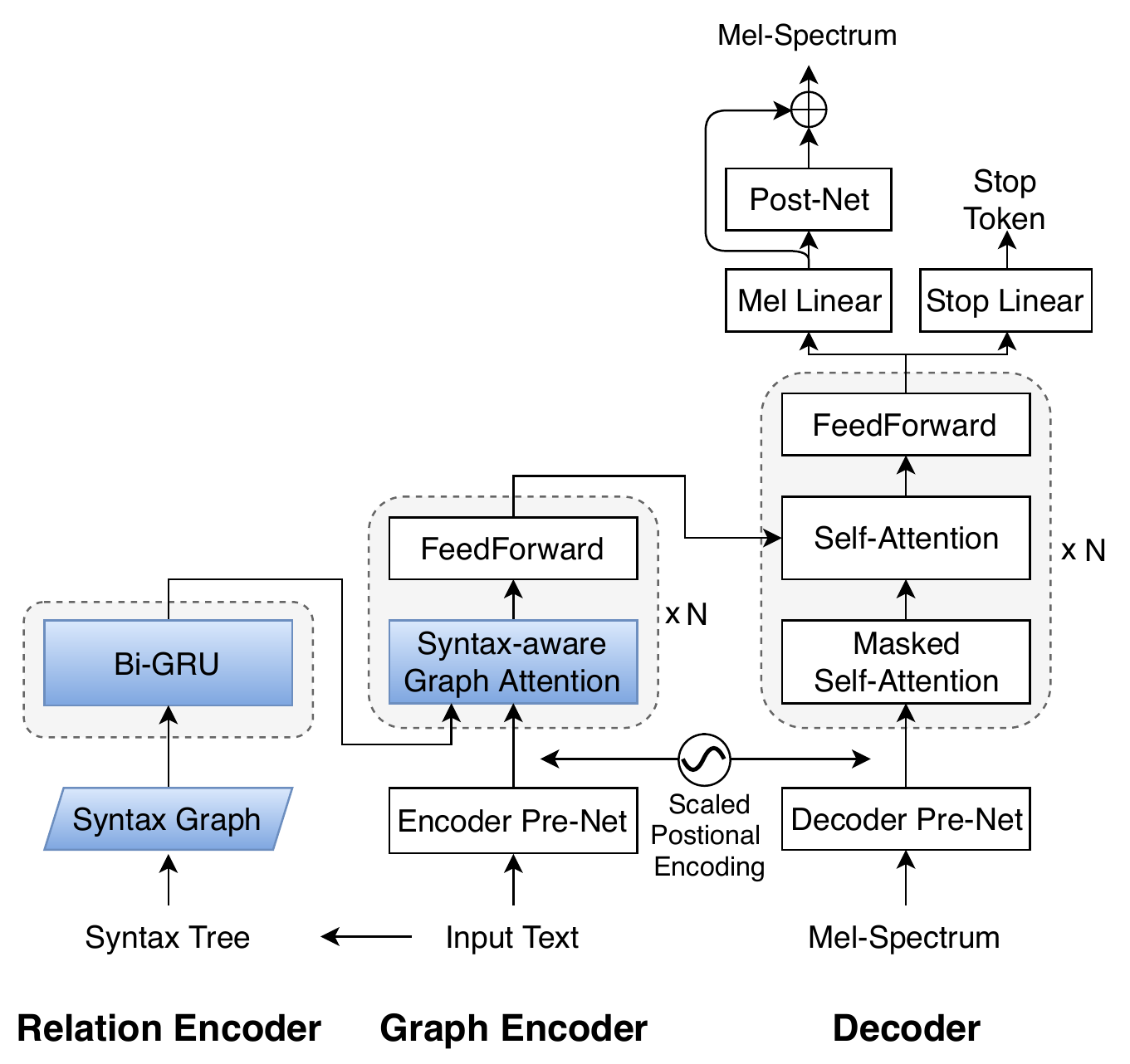}}
\vspace{-2mm}
\caption{The proposed \textit{GraphSpeech} that consists of Relation Encoder, Graph Encoder and Decoder. Its difference from Transformer TTS is highlighted in blue.}
\label{fig:model}
\vspace{-5mm}
\end{figure}

\vspace{-4mm}
\section{GraphSpeech}
\label{sec:model}
\vspace{-2mm}

Graphical structure plays an important role in natural language processing (NLP), and it often serves as the central formalism for representing syntax, semantics, and knowledge. We propose to incorporate graphical modeling into neural TTS architecture, called \textit{GraphSpeech} as shown Fig. \ref{fig:model}, to model the association between input text, its syntactic structure, and output speech utterances. 


 
In \textit{GraphSpeech}, the input to encoder includes both textual and syntactic knowledge.  It is known that the linguistic prosody of an utterance is closely related to the syntactic structure of a sentence. \textit{GraphSpeech} augments the speech synthesis input with such syntax information, in the form of syntax tree, that is expected to improve the linguistic representation of the input text. In practice, the syntax tree serves as an auxiliary signal for more accurate self-attention.  \textit{GraphSpeech} has a 3-step workflow: 1) \textit{Relation Encoder},  2) \textit{Graph Encoder} and 3) \textit{Decoder}, which will be discussed next. 

\vspace{-3mm}
\subsection{Relation Encoder}
\label{subsec:renc}
\vspace{-3mm}

Relation Encoder converts syntax tree of input text into a syntax graph that describes a global relationship among the involved input tokens. 
Syntactic dependency parse tree \cite{Nivre2007MaltParser,LiZP20} is one of the traditional ways to describe linguistic dependency relation among words. In a tree structure, only words that are directed related in a sentence are connected. Others have no direct connection. To excavate syntactic relation between two words in a sentence, we hope to extend the topological structure of the syntax tree to establish fully-connected communication.

\begin{figure}[tb]
\centering
\centerline{\includegraphics[width=1.03\linewidth]{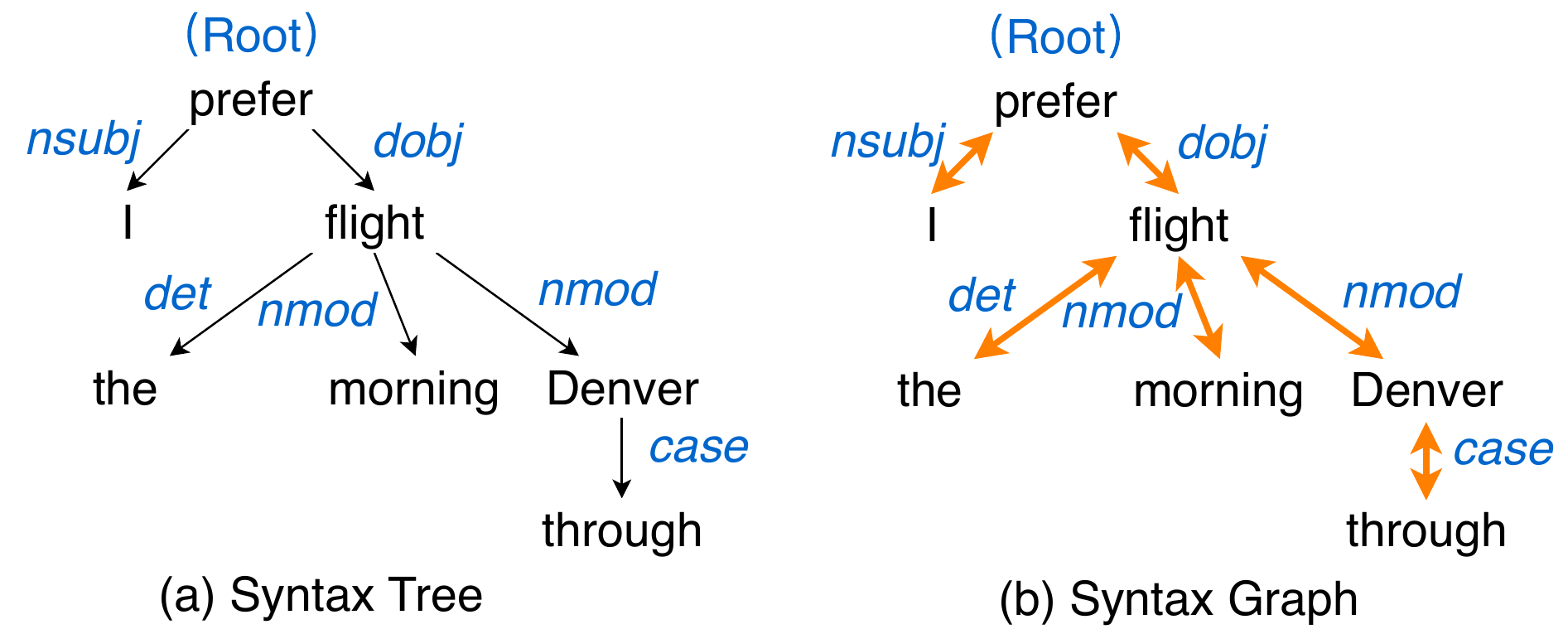}}
\vspace{-3mm}
\caption{An example of (a) syntax tree and (b) syntax graph of a sentence ``I prefer the morning flight through Denver.''. The blue words denote the dependency relation between two nodes. The black arrows denote directed one-way connections, and yellow arrows denote  two-way connections.  Self-loop connections are omitted due to space limitation.}
\label{fig:tree}
\vspace{-5mm}
\end{figure}

To this end, we propose \textit{syntax graph}, which is an extension of the syntax tree as shown in Fig. \ref{fig:tree}. Our idea is to change the one-way connection into a two-way connection by adding reverse connection. Moreover, we introduce self-loop edges with a specific label for each word. In this way, words in a sentence are represented by nodes, and their connections are denoted by edges. 
With the two-way connections as in in Fig. \ref{fig:tree}(b), a word is able to directly receive and send information to any other words whether they are directly connected or not.

To model the relations between two nodes, the relation between a node pair is depicted as the shortest relation path between them. We use recurrent neural networks with Gated Recurrent Unit (GRU) \cite{cho2014learning} to transform relation sequence into a distributed representation. The shortest relation path $sp_{i \rightarrow j}$ between the node $i$ and the node $j$ denoted as $ [sp_{1},..., sp_{t}, ..., sp_{n+1}] =  [e(i,k_{1}), e(k_{1}, k_{2}), ..., e(k_{n}, j)]$, where $e(\cdot, \cdot)$ indicates the edge label and $k_{1:n}$ are the relay nodes. We employ bi-directional GRUs for path sequence encoding: 
\begin{align}
\overrightarrow{s_{t}} & =\operatorname{GRU}_{f}\left(\overrightarrow{s_{t-1}}, s p_{t}\right)  \\
\overleftarrow{s_{t}} & =\operatorname{GRU}_{b}\left(\overleftarrow{s_{t+1}}, s p_{t}\right)
\end{align}

The last hidden states  of the forward GRU network and the backward GRU networks are concatenated to form the final relation encoding $r_{ij} = [\overrightarrow{s_{n+1}}; \overleftarrow{s_{0}}]$. The final relation encoding represents the linguistic relation between two words. 
In neural TTS, the basic unit in a sentence is character token. We extend the word-level relation encoding in NLP to character-level encoding here. If two characters belong to the same word, we use self-loop edge encoding $r_{ii}$ to define their relation encoding; if two characters belong to different words,  we directly use the $r_{ij}$, i.e. the relation encoding of the word they belong to, to assign their relation encoding. 

{The relation encoding
provides the model a global view about how information should be gathered and distributed, i.e., where to attend. Next we will discuss the proposed Graph Encoder, which is designed to incorporate the syntactic relation encoding inside a self-attention mechanism to indicate the character relationships.}



\vspace{-4mm}
\subsection{Graph Encoder}
\label{subsec:genc}
\vspace{-2mm}
Graph encoder aims to transform the input character embedding sequence and the relation encodings into a sequence of corresponding syntactically-motivated character embeddings. We incorporate explicit relation representation between two nodes into the attention calculation, denoted as syntax-aware graph attention.

Multiple blocks of syntax-aware graph attention and feed forward layer are then stacked to compute the final character representations. At each block, a character embedding is updated based on all other character embeddings and the corresponding relation encodings. The resulted character embeddings at the last block are fed to the decoder for acoustic feature generation.

Relation encoding only encodes the shortest path between two characters. To encode the direction of the connections when calculating attention, we first separate the relation encoding $r_{ij}$ into the forward relation encoding $r_{i \rightarrow j}$ and the backward relation encoding $r_{j \rightarrow i}$: $[r_{i \rightarrow j};r_{j \rightarrow i} ]=W_{r} r_{i j}$. Then, a syntax-aware graph attention was used to compute the attention score, which is based on both the character representations and their bi-directional relation representation:
 
\begin{equation}
\label{eq:att}
\begin{aligned}
s_{i j} &=g\left(x_{i}, x_{j}, r_{i j}\right) \\
&=\left(x_{i}+r_{i \rightarrow j}\right) W_{q}^{T} W_{k}\left(x_{j}+r_{j \rightarrow i}\right) \\
&=\underbrace{x_{i} W_{q}^{T} W_{k} x_{j}}_{(a)}+\underbrace{x_{i} W_{q}^{T} W_{k} r_{j \rightarrow i}}_{(b)} \\
&+\underbrace{r_{i \rightarrow j} W_{q}^{T} W_{k} x_{j}}_{(c)}+\underbrace{r_{i \rightarrow j} W_{q}^{T} W_{k} r_{j \rightarrow i}}_{(d)}
\end{aligned}
\end{equation}

We note that the terms in Eq. (\ref{eq:att}) is related to an intuitive interpretation by the formulation as follows: 1) term (a) captures purely content-based addressing, which is consistent with Eq. (\ref{eq:att0}); 2) term (b) represents a forward relation bias; 3) term (c) governs a backward relation bias; and 4) term (d) encodes the universal relation bias.


{With the syntax-aware graph attention, the syntax knowledge was used to guide the character encoding by incorporating explicit syntactic relation constraints into attention mechanism. Lastly, the syntactically-motivated character embeddings are taken by the decoder for acoustic feature prediction.}

\vspace{-4mm}
\subsection{Decoder}
\label{subsec:dec}
\vspace{-2mm}

The decoder follows the same structure reported in Transformer TTS \cite{2019transformerTTS}. We use Mel Linear, Post-Net and Stop Linear to predict the mel-spectrum and the stop token respectively. As discussed above, the relation encoder encodes the explicit syntactic relation between two distant characters. The graph encoder adopts the syntax-aware graph attention to derive syntactically-motivated character representations. The decoder takes the character representations as input and produces natural acoustic features. In this way, the decoder learns to associate input text, and its syntax, with output utterances.

\vspace{-5mm}
\section{Experiments}
\vspace{-3mm}
We conduct objective and subjective evaluations to assess the performance of our proposed GraphSpeech framework. We use LJSpeech database
~\cite{ljspeech} 
that consists of 13,100 short clips with a total of nearly 24 hours of speech from one single speaker reading about 7 non-fiction books. We use the state-of-the-art Transformer TTS \cite{2019transformerTTS} as the baseline.





\subsection{Experimental Setup}

In this study, we use Stanza \cite{qi2020stanza} to extract the syntactic dependency parser trees.
{For relation encoder, the size of edge embedding is set to 200-dimensional with random initialization. We set the size of GRU layer to 200 in both directions and generate the 200-dimensional relation encodings. The graph encoder takes 256-dimensional character embeddings, or node embeddings, as input.}
We use $N=6$ blocks in the graph encoder, and $N=6$ blocks in the decoder for  \textit{GraphSpeech}. 
We design 4 heads for multi-head attention in the graph encoder and decoder.
The decoder generates a sequence of 80-channel Mel-spectrum acoustic features as output. 

During training, we extract Mel-spectrum acoustic features with a frame size of 50ms and 12.5ms frame shift, that are further normalized to zero-mean and unit-variance, to serve as the reference target. We train the models using Adam optimizer with $\beta_1$ = 0.9, $\beta_2$ = 0.98. The same learning rate schedule as in~ \cite{vaswani2017attention} is adopted in our experiments.

For computational efficiency, we combine all the distinct shortest paths in a training and testing batch. We then encode them into vector representations by relation encoder, as described in Section \ref{subsec:renc}.
{For rapid turn-around, we use Griffin-Lim algorithm \cite{Griffin1984Signal} for waveform generation.}

\begin{table}[t]
\centering
\label{tabsss:mcd}
\vspace{-4mm}
\caption {The MCD, RMSE and MOS results of two systems in the comparison.}
\label{tab:mcd}
\begingroup
\begin{tabular}{p{3cm}p{2cm}p{2cm}}
\toprule
 \textbf{System}&  \textbf{MCD} [dB] & \textbf{RMSE} [Hz]  
 \\
\hline 
Transformer TTS & 8.021  & 1.782 
\\
\textbf{GraphSpeech} & \textbf{6.821} & \textbf{1.625}  
\\
\bottomrule
\end{tabular}
\vspace{-4mm}
\endgroup
\end{table}

\vspace{-4mm}
\subsection{Objective Evaluation}

We employ Mel-spectrum distortion (MCD) \cite{kubichek1993mel} to measure the spectral distance between the synthesized and reference Mel-spectrum features. We use Root Mean Squared Error (RMSE) as prosody evaluation metric \cite{sisman2018wavelet}. 

In Table \ref{tab:mcd}, we report the MCD and RMSE results in a comparative study. We observe that \textit{GraphSpeech} provides lower MCD and RMSE values that validate its effectiveness. With the syntax-aware graph attention mechanism,  \textit{GraphSpeech} learns to associate not only input text, but also its syntax structure with target utterances, that improves TTS speech rendering.

\vspace{-4mm}
\subsection{Subjective Evaluation}
\vspace{-1mm}
 
We conduct listening experiments for subjective evaluation. We first evaluate Transformer TTS, \textit{GraphSpeech} and ground truth in terms of mean opinion score (MOS). 
The listeners grade the utterances at a 5-point scale: ``5'' for excellent, ``4'' for good, ``3'' for fair, ``2'' for poor, and ``1'' for bad.
15 subjects participate in these experiments, each listening to 100 synthesized speech samples.

\begin{figure}[tb]
\centering
\centerline{\includegraphics[width=0.7\linewidth]{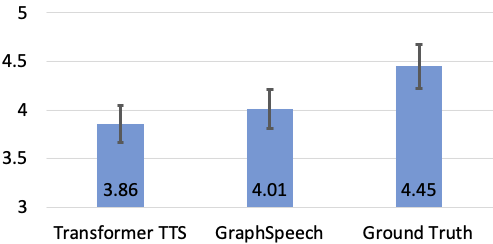}}
\vspace{-3mm}
\caption{The MOS results of two systems in the comparison with 95\% confidence interval.}
\label{fig:mos}
\vspace{-1mm}
\end{figure}

\begin{figure}[tb]
\centering
\centerline{\includegraphics[width=\linewidth]{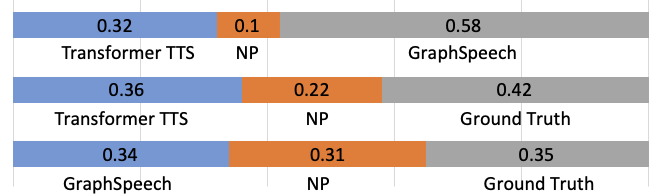}}
\vspace{-3mm}
\caption{The AB preference test for naturalness evaluation
by 15 listeners.
The $p$-values are $2.0012e^{\--20}$, $1.9234e^{\--20}$ and $2.0182e^{\--20}$ respectively.}
\label{fig:ab}
\vspace{-5mm}
\end{figure}

As shown in Fig.~\ref{fig:mos}, GraphSpeech outperforms Transformer TTS by a large margin, and achieves comparable results to that of ground truth natural speech, which we believe is remarkable. These results clearly show that the syntax-aware graph attention strategy, as proposed and implemented in \textit{GraphSpeech}, effectively model the linguistic knowledge and achieves high-quality synthesized voice. 

We further conduct AB preference tests where the listeners are asked to compare the quality and naturalness of the synthesized speech samples between a pair of systems, and select the better one. As can be seen in Fig. \ref{fig:ab}, listeners prefer the proposed \textit{GraphSpeech} model consistently, which further validates the proposed graph attention mechanism. It is reported that the samples synthesized by \textit{GraphSpeech} is close to natural speech, which is encouraging.

\vspace{-3mm}
\section{Conclusion}
\vspace{-3mm}

In this work, we propose a novel neural TTS architecture, denoted as \textit{GraphSpeech}. The proposed syntax-aware graph attention mechanism effectively models the linguistic relation between any two characters of the input text. Experimental results show that \textit{GraphSpeech} outperforms the original self-attention systems in both objective and subjective evaluations, and achieves remarkable performance in terms of voice quality and prosody naturalness.

\vspace{-2mm}
\section{Acknowledgements}
\vspace{-2mm}
The research by Rui Liu and Berrak Sisman is funded by SUTD Start-up Grant Artificial Intelligence for Human Voice Conversion (SRG ISTD 2020 158) and SUTD AI Grant, titled `The Understanding and Synthesis of Expressive Speech by AI'.
This research by Haizhou Li is supported by the National Research Foundation, Singapore under its AI Singapore Programme (Award No: AISG-GC-2019-002) and (Award No: AISG-100E-2018-006), and its National Robotics Programme (Grant No. 192 25 00054), and by RIE2020 Advanced Manufacturing and Engineering Programmatic Grants A1687b0033, and A18A2b0046. 
 



\bibliographystyle{IEEEbib}
{\footnotesize
\bibliography{strings}}
\end{document}